\pdfoutput=1

\documentclass[11pt]{article}
\usepackage{acl}

\usepackage{times}
\usepackage{latexsym}

\usepackage[T1]{fontenc}

\usepackage[utf8]{inputenc}

\usepackage{microtype}

\usepackage{inconsolata}
\usepackage{graphicx}
\usepackage{booktabs}
\usepackage{xcolor}
\usepackage{amsmath}
\usepackage{algorithm}
\usepackage{algorithmic}
\usepackage{graphicx}
\usepackage{subfigure}
\usepackage{xfrac}
\usepackage{colortbl}
\usepackage{booktabs}  
\usepackage{amsfonts}  
\usepackage{nicefrac}
\usepackage{latexsym}
\usepackage{multirow}
\usepackage{amsmath}
\usepackage{tabularx}
\usepackage{makecell}
\usepackage{wrapfig}
\usepackage{enumerate}
\usepackage{engord}
\usepackage{hyperref}

\usepackage{tcolorbox}
\usepackage{xcolor}   
\tcbuselibrary{theorems}
\newtcbtheorem[number within=section]{mytheo}{My Theorem}%
{colback=green!5,colframe=green!35!black,fonttitle=\bfseries}{th}
\definecolor{change}{rgb}{1,0,0}  

\usepackage{orcidlink}
\usepackage{authblk}

\usepackage{xspace}
\def\ours{\textsc{P$^2$G}\xspace}
\def\ourswb{\textsc{P$^2$G}\xspace}

\def\oursBenchmark{\textsc{P$^2$GB}\xspace}
\def\oursBenchmarkwb{\textsc{P$^2$GB}\xspace}

\begin{document}

\title{Plug-and-Play Grounding of Reasoning in Multimodal Large \\Language Models} 

\author{
\textbf{Jiaxing Chen}$^{\dagger}$\textsuperscript{1}\thanks{Work done during Jiaxing and Dehu's internship at DeepGlint. $\dagger$: Equal contribution.}, 
\textbf{Yuxuan Liu}$^{\dagger}$\textsuperscript{1},
\textbf{Dehu Li}\textsuperscript{1} \\ 
\textbf{Xiang An}\textsuperscript{2}, 
\textbf{Weimo Deng}\textsuperscript{2}, 
\textbf{Ziyong Feng}\textsuperscript{2}, 
\textbf{Yongle Zhao}\textsuperscript{2},
\textbf{Yin Xie}\textsuperscript{2}    
\vspace{3px}
\\
\textsuperscript{1}Peking University
\textsuperscript{2}DeepGlint \\
\small{\texttt{\{jiaxing.chen, yx.liu, dehuli\}@stu.pku.edu.cn}}
}
\maketitle

\begin{abstract}
The rise of Multimodal Large Language Models (MLLMs), renowned for their advanced instruction-following and reasoning capabilities, has significantly propelled the field of visual reasoning. However, due to limitations in their image tokenization processes, most MLLMs struggle to capture fine details of text and objects in images, especially in high-resolution samples. To overcome this limitation, we introduce \ours, a novel framework for plug-and-play grounding in MLLMs. \ours utilizes the tool-usage potential of MLLMs to employ expert agents for on-the-fly grounding of reasoning into critical visual and textual elements in images, thereby enabling deliberate reasoning through multimodal prompting. Additionally, we develop \oursBenchmark, a benchmark designed to evaluate MLLMs' proficiency in understanding inter-object relationships and textual content in challenging high-resolution images. Extensive experiments on visual reasoning tasks demonstrate the superiority of \ours, achieving performance comparable to GPT-4V on \oursBenchmark with a 7B backbone. Our work underscores the potential of grounding reasoning with external agents in MLLMs, presenting a promising alternative to mere model scaling.
\end{abstract}

\section{Introduction}
\label{sec:intro}

Large language models (LLMs) \cite{touvron2023llama, Achiam2023GPT4TR, touvron2023llama2} have shown strong potential as a unified backbone for various language tasks, including in-context learning \cite{brown2020language, wang2022self}, instruction following \cite{ouyang2022training}, and reasoning \cite{sun2023survey, wang2023demonemnlp}. 

Extending LLMs to multimodal capabilities, researchers have developed Multimodal Large Language Models (MLLMs) \cite{zhu2023minigpt, liu2024visual-llava1, huang2024language, alayrac2022flamingo, wang2023cogvlm, instructblip}, treating each modality as a foreign language \cite{huang2024language, wu2023next}. These MLLMs show significant results in the field of visual reasoning.

Despite these advancements, MLLMs face limitations in visual reasoning due to the high demand for large-scale annotated data for vision instruction tuning \cite{zhu2023minigpt, liu2024visual-llava1}. Collecting annotated multimedia training examples is challenging, and multimodal instruction tuning data is even harder to scale. Another limitation is capturing details in high-resolution images or those with complex textual information, often leading to hallucinations or incorrect reasoning. Non-lossless tokenization of images can also overlook critical semantic details.

To address these challenges, successor works have explored grounding reasoning in MLLMs. \textsc{Kosmos-2} \cite{peng2023kosmos} and CogVLM \cite{wang2023cogvlm} generate bounding boxes for visual occurrences. LLaVAR \cite{zhang2023llavar} and TGDoc \cite{wang2023towardstgdoc} augment instruction tuning data with OCR-based textual clues and bounding boxes. However, these methods require large amounts of data and training costs.

Inspired by recent studies showing LLMs' effective use of external tools and agents \cite{shen2024hugginggpt, zhuang2024toolqa}, we propose \ours, a novel framework for plug-and-play grounding of reasoning in MLLMs. Instead of training MLLMs from scratch, we leverage lightweight proxy models as agents to obtain critical clues for reasoning. We use an OCR agent (via PaddleOCR \cite{paddleocr}) and a visual grounding agent (via Grounding-DINO \cite{liu2023grounding}) for text-rich and high-definition images. MLLMs generate specific queries for these agents based on the complexity of the reasoning task.

To evaluate \ourswb, we introduce \oursBenchmark, a challenging Visual Question Answering (VQA) benchmark designed to assess MLLMs' visual grounding, especially in high-resolution and text-rich scenarios. Our experiments on visual reasoning tasks, including \oursBenchmarkwb, demonstrate the superiority of \ours. Notably, \ourswb achieved comparable performance to GPT-4V on \oursBenchmark with a 7B backbone. Our work highlights the potential of plug-and-play grounding of reasoning as an alternative to model scaling. Our contributions are three-fold:

\begin{enumerate}[1)]
  \item We propose \ours, a framework for plug-and-play grounding of reasoning in high-resolution and text-rich visual scenarios using agents.
  \item We introduce \oursBenchmark, a VQA benchmark to assess MLLMs' reasoning capability in text-rich and high-definition image queries.
  \item We conduct extensive experiments on challenging reasoning datasets, demonstrating \ours's superior performance with a 7B MLLM backbone, surpassing similarly scaled or larger models.
\end{enumerate}

\begin{figure*}[htbp]
  \centering
  \includegraphics[width=\textwidth]{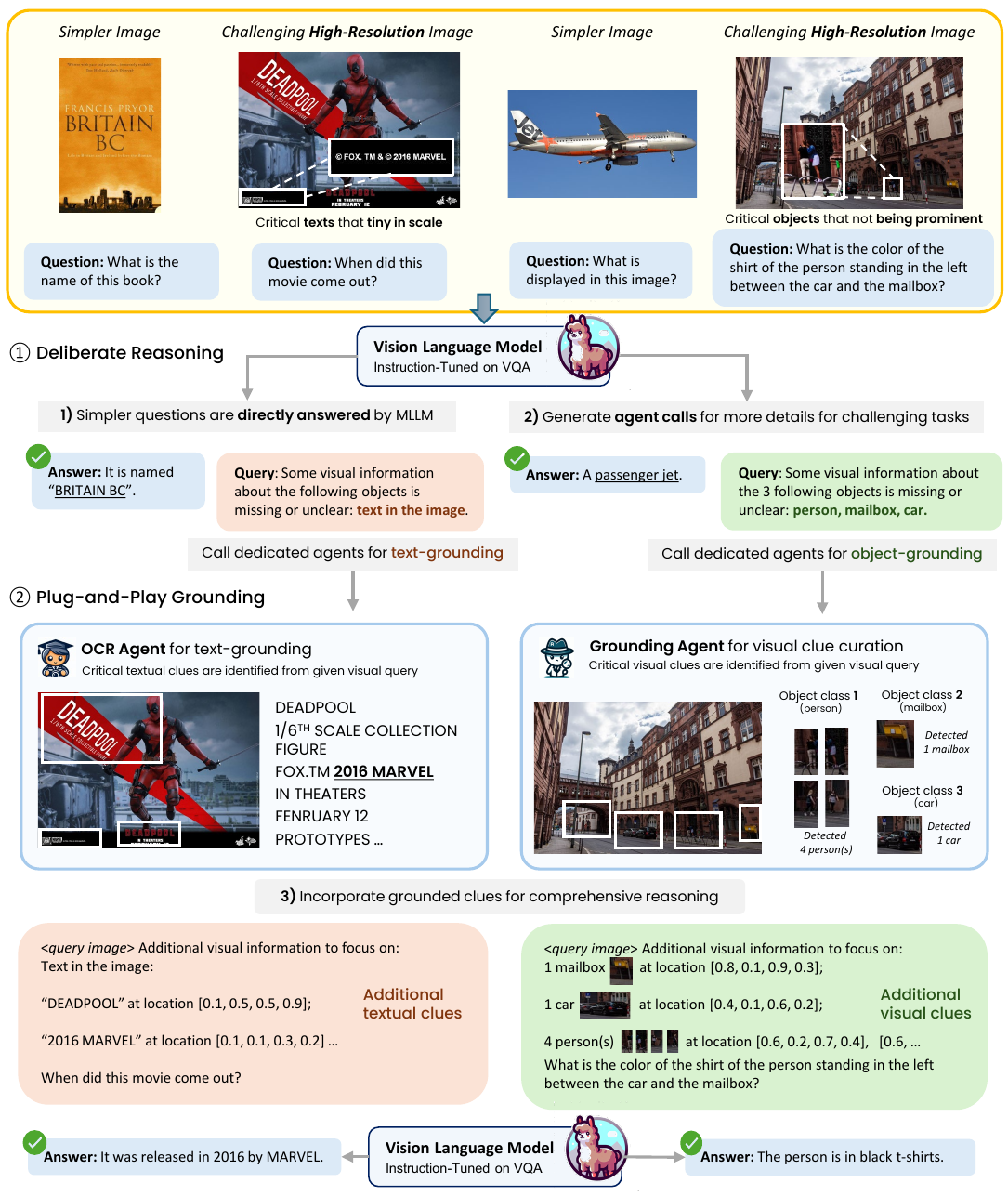}
  \vspace{5px}
  \caption{Illustration of our proposed \ourswb for grounding visual reasoning. Given a multi-modal query including an image and its corresponding question, (1) \ourswb first deliberately decide whether to seek additional clues (anticipated text and/or visual objects) from dedicated textual and/or visual grounding agents, or provide a direct answer for simple and confident cases. For challenging cases, (2) additional text or visual clues are then obtained via OCR Agent (\textit{text}) or Grounding Agent (\textit{image}) according to MLLM's request. Specifically, we include OCR texts and their relative positions for textual clues, and for visual clues, we detect and locate all objects for each requested class. Finally, we incorporate these clues into a multi-modal prompt for obtaining a grounded reasoning answer.}
  \label{fig:main_method}
\end{figure*}

\section{Methods}
\label{sec:method}

Our proposed framework, which we refer to as \ours, primarily addresses the challenge of visual reasoning tasks that involve high-resolution natural images and text-rich images. Our goal is to enhance the model's ability to interpret and analyze these complex visual inputs effectively, thereby improving its performance on visual reasoning that requires a nuanced understanding of both visual and textual elements in detail.

\subsection{Overall Design of \ours}
Figure \ref{fig:main_method} illustrates the proposed \ours: \textbf{P}lug-and-\textbf{P}lay \textbf{G}rounding of Reasoning in large vision language models. The key objective of \ourswb lies in enhancing the groundedness and factualness of reasoning from multimodal language models (MLLMs), without relying on heavily supervised (instruction) fine-tuning on extensive annotated data. And to achieve this objective, we harness the emergent capabilities like \textit{in-context learning} \cite{dong2022survey}, \textit{instruction following} \cite{longpre2023flan} and \textit{tool-usage} \cite{shen2024hugginggpt} capability of large language models. Below, we introduce the procedure of \ourswb in detail.

\subsubsection{Deliberate Reasoning} To ground the reasoning procedure of MLLMs, one key challenge is the hallucination of reasoning paths. In other words, MLLMs must know their \textit{don't-knows} \cite{cheng2024can} ahead. To mitigate this issue, we propose Deliberate Reasoning in \ours, which encourages the MLLMs to first assess their current ability to solve the provided question, before moving forward on reasoning. 

As illustrated in Figure \ref{fig:main_method}, for a simple visual query, \ourswb generates the correct answer directly, while for challenging cases, \ourswb autonomously assesses its current capability, and poses demand on support from external agents (experts) on specific textual or visual supporting clues (in the form of natural language query). By introducing this \textit{deliberate reasoning} process before moving on to the reasoning problem, we could thereby empower the MLLM with external agents for concise textual or visual understanding, which is generally challenging for large vision language models, especially for nuanced but important details high-definition images. The capability of deliberate reasoning ahead is attained through dedicated instruction tuning, which we will elaborate on in Sec. \ref{sec:inst}.

\subsubsection{Plug-and-Play Grounding} The surging works in the field of retrieval augmented generation (RAG) \cite{gao2023retrieval} and tool-usage \cite{shen2024hugginggpt, liang2023taskmatrix} inspired us on leveraging external experts (agents) in grounding multimodal reasoning with rich textual and visual facts and clues. One major challenge for MLLMs in reasoning \cite{liu2023improved, liu2024visual-llava1, ye2023mplug} is the expressiveness of image representation, where an \textit{only representation} (visual tokens) is provided for reasoning, which hinders the comprehensiveness of encompassed visual information, especially under high-definition or text-rich scenarios. The information loss during such auto-encoding compression refrains MLLM from generating grounded, accurate reasoning. The latest works either fine-tune on more VQA data \cite{zhang2023llavar}, or prepend OCR texts into context \cite{wang2023towardstgdoc, liu2023largebringocr}, which does not essentially mitigate this core limitation.

As a step forward, we propose \textit{Plug-and-Play} Grounding in \ours, to mitigate the limitation above by providing both rich textual and visual clues, leveraging external agents (experts). As illustrated in Figure \ref{fig:main_method}, based on the specific query on semantic details from MLLMs, we correspondingly call \textit{1) OCR Agent} to collect text pieces, or \textit{2) Grounding Agent} to fetch visual patches corresponding to the crucial semantic objects requested by the MLLM. Beyond fetching these semantic premises, we also incorporate their relevant position in the image into a multi-modal question prompt, before obtaining a final comprehensive reasoning answer. Such plug-and-play design enables us to leverage SOTA text (PaddleOCR \cite{paddleocr}) or image (Grounding DINO \cite{liu2023grounding}) processing tools, mitigating the demand for dedicated tuning of backbone MLLMs. By providing dedicated textual and visual clues, we significantly improve the correctness and groundedness of MLLM's reasoning. Details are described in Sec. \ref{sec:modeling}.

\subsection{Model Structure}
\label{sec:modeling}
\subsubsection{Architectural Designs}
\ours integrates four main components: an LLM, a vision encoder, a projection module, and textual (OCR) and visual grounding agents. These components work jointly to enhance the model's ability to process and interpret complex multimodal data.

We use Vicuna-7B-V1.3 \cite{zheng2024judging} as our LLM, which trained Llama on approximately 125K conversations collected from ShareGPT.com.
The vision encoder is CLIP ViT-L/14, which processes inputs resized and padded to \(224^2\). This encoder handles both the original images and specific regions containing detected objects.

To map visual semantics to the LLM's hidden space, we use two types of projection modules: an MLP module, and a cross-attention-based Resampler \cite{alayrac2022flamingo}. The MLP maintains the count of visual tokens and only reshapes its dimension, while the Resampler (one-layer cross attention)\footnote{The resampler is implemented as a single-layer cross-attention, following \citet{alayrac2022flamingo}.} also reduces the token quantity (from 256 to 32) to ensure an efficient context.

To maintain an adequate count of visual tokens, we toggle between the two projection modules. For inputs with only initial (global) image features, the MLP maps all visual tokens. 
For inputs with 1 to 4 critical objects, we employ an MLP to map the visual features of these objects and utilize the Resampler to downsample the global image.
When more than 4 objects are detected via Grounding Agent, the Resampler handles all visual features of objects to ensure an efficient context size.

The Grounding Agent uses Grounding DINO \cite{liu2023grounding} to identify and extract relevant objects, while the OCR Agent utilizes PaddleOCR\footnote{https://github.com/PaddlePaddle/PaddleOCR} to retrieve textual information.

\subsubsection{Deliberate Reasoning and Plug-and-Play Grounding}
We detail the plug-and-play grounding of reasoning in \ours. As shown in Figure \ref{fig:main_method}, the model first determines if additional visual or textual clues are needed. For straightforward ones, the model directly outputs its reasoning. For high-resolution images or those with detailed text, the model generates query responses, calling the OCR or Grounding Agent. Such capability is attained through instruction fine-tuning, detailed in Section \ref{sec:inst}.

\begin{table}[t]
\centering
\resizebox{\linewidth}{!}{
\begin{tcolorbox}[width=0.5\textwidth, fontupper=\small, colback=blue!2, boxrule=0.9pt]
Sorry, I cannot answer the question. Some visual information about the following objects is missing or unclear: \textcolor{purple}{object\textsubscript{1}, \dots, object\textsubscript{n}}.
\end{tcolorbox}}
\vspace{-15px}
\captionof{figure}{Calling \textit{Grounding Agent} for visual clues.}
\label{tab:prompt1} 
\end{table}

\begin{table}[t]
\centering
\resizebox{\linewidth}{!}{
\begin{tcolorbox}[width=0.5\textwidth, fontupper=\small, colback=blue!2, boxrule=0.9pt] 
$<$image$>$ \textcolor{purple}{(Original image)}\\

Additional visual information to focus on: \\
3 button(s) $<$object$>$, $<$object$>$, $<$object$>$ at location [0.25, 0.63, 0.26, 0.64], [0.47, 0.59, 0.48, 0.60], [0.52, 0.62, 0.53, 0.63] \\

1 paper clip $<$object$>$ at location [0.65, 0.70, 0.66 ... \\
\textcolor{blue}{(Object features and their positions)}\\

[object class] not existent in the image ... \textcolor{violet}{(Objects that not detected by Grounding Agents)} \\

Are all buttons in the image larger than the paper clips? \\
Answer the question using a single word or phrase. \textcolor{purple}{(Original question)}
\end{tcolorbox}}
\vspace{-10px}
\captionof{figure}{Example prompt for the model's second round of reasoning, with visual clues from \textit{Grounding Agent}.}
\label{tab:example1} 
\end{table}

For high-resolution images, the model's initial response may miss certain objects or details, as shown in Figure~\ref{tab:prompt1}. Grounding DINO detects and crops these objects, magnifying them for focused analysis. These crops are incorporated into prompts for a second round of inference, as illustrated in Figure~\ref{tab:example1}, enabling the model to provide more accurate answers.
This process is formalized with a detection function \(F_d\), which processes an image \(I\) and a set of target objects \( \{object_1, \dots, object_n\} \), resulting in image crops \(P\):
\begin{equation}
P = F_d(I, \{object_1, \dots, object_n\}),
\end{equation}
where \(P = \{p_1, p_2, \ldots, p_m\}\) are the image crops identified by Grounding DINO. The total number of objects and individual quantities of each type are related by \(\sum_{i=1}^{n} x_i = m\), where \(n\) is the total number of object types and \(x_i\) is the quantity of the \(i\)-th object. As illustrated in Figure~\ref{tab:example1}, we also inform MLLMs of the objects not being detected, indicating their potential absence from the image.

\begin{table}[t]
\centering
\resizebox{\linewidth}{!}{
\begin{tcolorbox}[width=0.5\textwidth, fontupper=\small, colback=blue!2, boxrule=0.9pt] 
Sorry, I cannot answer the question. Some visual information about the following objects is missing or unclear: \textcolor{purple}{text in the image}.
\end{tcolorbox}}
\vspace{-15px}
\captionof{figure}{Calling \textit{OCR Agent} for textual clues.}
\label{tab:prompt2} 
\end{table}

\begin{table}[t]
\centering
\resizebox{\linewidth}{!}{
\begin{tcolorbox}[width=0.5\textwidth, fontupper=\small, colback=blue!2, boxrule=0.9pt] 

$<$image$>$ \textcolor{purple}{(Original image)}\\ 

Additional visual information to focus on:\\
Text in the image: `May311918' at location [0.66, 0.043, 0.931, 0.077]; `3379Bark Jane Rd' at location [0.545, 0.103, 0.921, 0.131]. \textcolor{blue}{(Text and their positions)}\\

\textcolor{brown}{
Please focus on providing an answer to the question without considering any challenges related to the clarity or presence of text in the image.}\\
\textcolor{violet}{(Add this segment when no text detected in image)}\\

By whom is this letter written? \textcolor{purple}{(Original question)}

\end{tcolorbox}}
\vspace{-15px}
\captionof{figure}{Example prompt for the model's second round of reasoning with textual clues from \textit{OCR Agent}.}
\label{tab:example2} 
\end{table}

For text-rich images, the model's call to the OCR Agent is shown in Figure~\ref{tab:prompt2}. PaddleOCR extracts textual elements, which are integrated with bounding boxes and questions, as shown in Figure~\ref{tab:example2}. This enhances the model's recognition of text presence and positions.
Given additional textual clues $\mathcal{T}$ and visual clues $\mathcal{P}$ from external agents, we obtain the final visual reasoning results via:
\begin{equation}
    \mathcal{R} = \text{MLLM}(q_i, q_t, \mathcal{T}, \mathcal{P}),
    \label{eq:inf}
\end{equation}
where $q_i$ and $q_t$ denote image and text queries, respectively. By conditioning on both image $q_i$ and enriched information $\mathcal{T}$ and $\mathcal{P}$, we achieve plug-and-play grounding of reasoning, leveraging MLLMs' in-context learning and instruction-following capabilities.

\subsection{Training of \ours}
\label{sec:inst}
We outline the training process to equip \ours with multimodal capabilities and deliberate reasoning. It consists of two stages: multimodal instruction tuning and learning of deliberate reasoning, each designed to progressively build the \ours's ability to handle complex visual and textual inputs.

\subsubsection{Multimodal Instruction Tuning}
The first stage focuses on equipping our base LLM (Vicuna-7B-V1.3 \cite{zheng2024judging}) with fundamental multimodal capabilities. We follow the procedures established in LLaVA \cite{liu2024visual-llava1}. We employ a 80K sample from LLaVA instruction data, following the procedures and splits used in V*~\cite{wu2023v}. This stage brings fundamental multimodal capabilities to LLMs.

\subsubsection{Learning of Deliberate Reasoning}
Our second stage aims to refine \ours's ability to reason deliberately, using agents to gather additional clues when needed. It involves two key steps: (1) \textit{Identifying Need for Additional Information}. The model learns to differentiate between straightforward and complex queries: Simple queries are answered directly, while complex queries trigger the use of OCR and grounding agents to gather additional textual or visual information.
(2) \textit{Learning to incorporate Additional Information}. We curate a set of challenging VQA queries, consisting of both \textit{positive} and \textit{negative} samples. Negative samples train the model to recognize its deficiency and generate agent calls. Positive samples (including both straightforward and complex queries) help the model to utilize additional clues from agents effectively.

Particularly, we adopt a two-round approach: the first stage for direct answering or generating agent calls (\textit{round 1}), and the second stage for utilizing multimodal clues (\textit{round 2}).
(1) For \textbf{text-rich image reasoning}, we select data from train sets of ChartVQA, DOCVQA, and TextVQA, focusing on images with resolutions over 500 pixels and critical texts smaller than 20 pixels. We pre-extract texts with PaddleOCR. The data was then split into negative samples (indicating the need for additional text) and positive samples.

(2) For \textbf{visual object grounding}, we adapt data from V* ~\cite{wu2023v} to improve the model's understanding of quantitative relationships and spatial arrangements between objects by incorporating the number of objects and their bounding boxes.
Our two-stage training process ensures \ours handle both simple and complex multimodal queries, leveraging additional information when necessary to provide accurate, grounded answers.

\begin{figure*}[t]
  \centering
  \includegraphics[width=\textwidth]{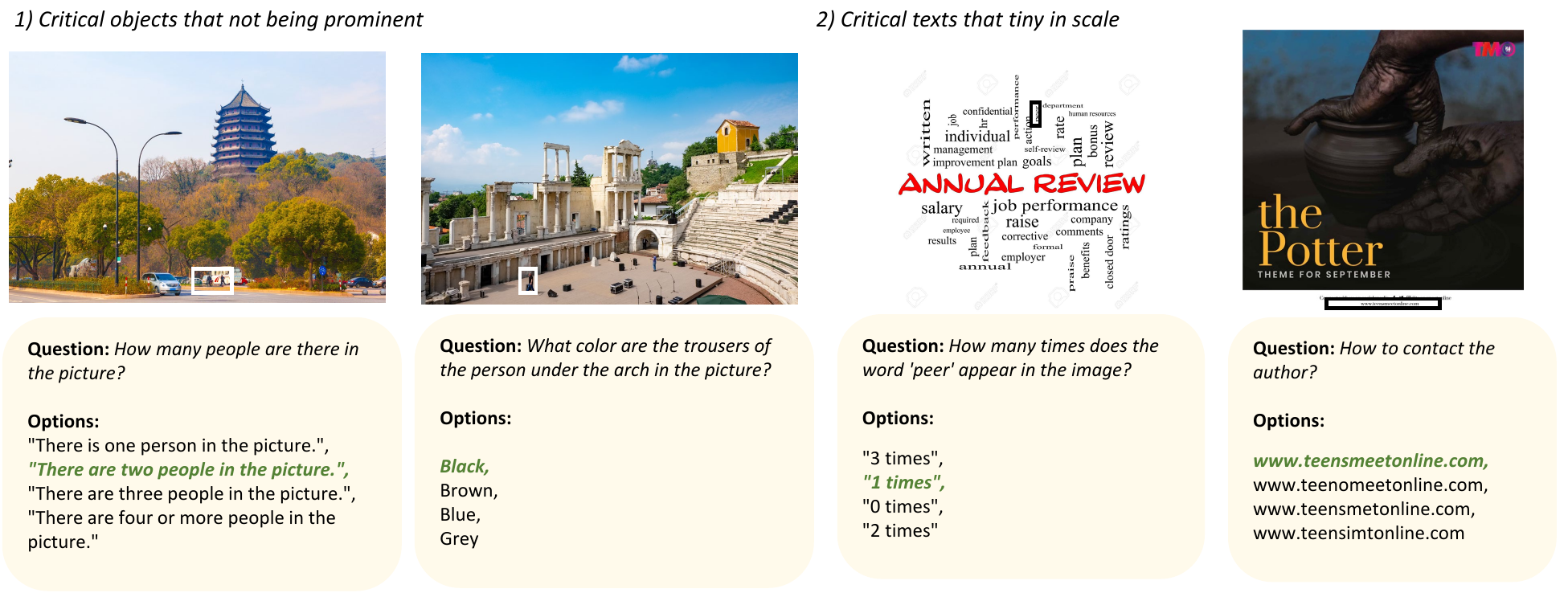}
  \caption{Illustration of our proposed \oursBenchmarkwb benchmark. In \oursBenchmark, we consider two challenging visual reasoning scenarios: comprehensive image understanding and text-rich visual reasoning. For the former, we delicately collect high-definition image samples where the critical object is not prominent (i.e., tiny in scale) and challenging to identify, while for the latter we include samples in which crucial textual parts are tiny as well.}
  \label{fig:main_benchmark}
\end{figure*}

\section{\oursBenchmarkwb Benchmark}

To quantitatively assess the visual reasoning capabilities under text-rich or high-resolution scenarios, we constructed a challenging benchmark \oursBenchmark. It includes Comprehensive Image Understanding with Fine-grained Recognition (2080 samples) and Image Text Content Understanding (50 samples), totaling 2130 samples (pair of an image and multiple-choice question)\footnote{The proposed benchmark will be released publicly.}.

(1) \textit{Comprehensive Image Understanding with Fine-grained Recognition} involves analysing high-resolution images with complex scenes containing multiple objects that the model must identify and describe, including their types, locations, and interactions, to test its ability to recognize and distinguish objects within the scene.
For this task, we randomly select images from SA-1B \citep{iccvsam1b} dataset and adopt EVA-02-L \citep{eva02l} detector to extract small object (detection boxes) from the images. For each image, the top 5 boxes are retained based on their scores. A detection box is considered a small object if its area is less than 1/10 of the full image. We use GPT-4o as a candidate for generating questions for each image. In each image, a red visual box is used to mark the object that needs to be questioned. GPT-4o generates a question based on the red box, with four answer options and one correct answer. The questions, options, and answers are all manually reviewed subsequently for accuracy, clarity, and does not contain biased or toxic contents.

(2) \textit{Image Text Content Understanding} involves identifying and understanding small textual content within high-resolution images and answering related questions. This task tests the model's ability to discern fine text and engage in logical reasoning based on the text. As illustrated in Figure \ref{fig:main_benchmark}, we design multiple-choice answers for each question that carefully crafted and manually reviewed to ensure validity, fairness, and eliminate ambiguities. To construct this benchmark, we adapt the PowerPoint images and questions from \citep{wang2023towardstgdoc}, and manually select challenging samples that wider than 1,000 pixels, contains tiny crucial texts, and paired with difficult questions.

\section{Experiments}
\subsection{Experimental Setup}

\begin{table*}[!t]
\tabcolsep 0.15 in
\centering

\resizebox{\textwidth}{!}{
\begin{tabular}{c|c|p{1.3cm}<{\centering}p{1.3cm}<{\centering}p{1.2cm}<{\centering}p{1.2cm}<{\centering}p{1.2cm}<{\centering}p{1.3cm}<{\centering}}
\toprule
Model & Size & DocVQA & ChartVQA & GQA & SEED & MMVET & MME \\ 
\midrule
MiniGPT-4 \cite{zhu2023minigpt} & 7B & 3.0 & 4.3 & - & - & - & - \\ 
mPLUG-OWL \cite{ye2023mplug} & 7B & 6.9 & 9.5 & - & - & - & - \\ 
LlaVAR \cite{zhang2023llavar} & 7B & 11.6 & 8.0 & - & - & - & - \\ 
TGDoc \cite{wang2023towardstgdoc} & 7B & 9.0 & 12.72  & - & - & - & - \\
LLaVA \cite{liu2024visual-llava1} & 7B & 19.06 & 15.30 & 17.09 & 23.50 & 29.10 & 1107 \\ 
Instruct-BLIP \cite{instructblip} & 7B & - & - & 49.20 & - & 26.20 & - \\ 
LLaVA \cite{liu2024visual-llava1} & 13B & 31.77 & 25.70 & 17.09 & 24.01 & 32.70 & 965 \\ 
Instruct-BLIP \cite{instructblip} & 13B & - & - & 49.50 & - & 25.60 & - \\ 
\midrule
LLaVA + \ourswb (Ours) & 7B & \textbf{61.44} & \textbf{37.20} & \textbf{59.87} & \textbf{27.46} & \textbf{32.90} & \textbf{1223} \\ 
\bottomrule
\end{tabular}
}
\caption{Performance of \ourswb on visual reasoning tasks. The best performing 7B model is marked in \textbf{bold}.}
\label{tab:main_tab}
\end{table*}

\begin{table}[!t]
\centering
\resizebox{\linewidth}{!}{
\begin{tabular}{c|c|p{1.5cm}<{\centering}p{1.3cm}<{\centering}p{1.3cm}}
\toprule
Model & Size & Objects & Texts \\
\midrule
GPT-4V \cite{Achiam2023GPT4TR} & $>$1T & \underline{50.1} & \underline{68.0} \\
LLaVA (Vicuna-1.3) & 7B & 40.1 & 8.0 \\
LLaVA (Vicuna-1.3) & 13B & 40.2 & 8.0 \\
\midrule
LLaVA + \ourswb (Ours) & 7B & \textbf{42.5} & \textbf{50.0} \\
Gain (\%) & - & \textcolor{blue}{1.06$\times$} & \textcolor{blue}{6.3$\times$} \\
\bottomrule
\end{tabular}}
\caption{Experimental results of \ours and baselines on our challenging high-resolution benchmark \oursBenchmark.}
\vspace{-10px}
\label{tab:main_tab}
\end{table}

\paragraph{Models and Baselines} For MLLMs, we select Vicuna-7B-V1.3 \cite{chiang2023vicuna} as the language backbone, and follow LLaVA to train an MLLM backbone for \ours. To build up two agents for visual and textual grounding, we select Grounding DINO \cite{liu2023grounding} for obtaining visual clues (i.e., objects) and PaddleOCR \cite{paddleocr} for screening texts within the image query. We compare \ourswb against multiple similar-scaled, instruction-tuned MLLMs, including vanilla LLaVA \cite{liu2024visual-llava1}, MiniGPT-4 \cite{zhu2023minigpt}, mPLUG-OWL \cite{ye2023mplug}, and Instruct-BLIP \cite{instructblip}. In addition, we compare \ourswb against MLLMs dedicated optimized for semantic-rich reasoning, i.e., LLaVAR \cite{zhang2023llavar}, and TGDoc \cite{wang2023towardstgdoc}. Finally, we include the most capable MLLM so far, GPT-4V \cite{Achiam2023GPT4TR} on our challenging benchmark \oursBenchmark.

\paragraph{Datasets} Following previous works, we test \ourswb on a variety of visual reasoning benchmarks. For text-rich visual reasoning, we select DocVQA \cite{mathew2021docvqa} and ChartVQA \cite{masry2022chartqa}, and GQA \cite{hudson2019gqa}, SEED \cite{li2023seed}, MM-VET \cite{yu2023mmvet}, and MME \cite{li2023seed} for semantic-rich and general visual reasoning. Beyond existing benchmarks, we also curate a challenging benchmark \oursBenchmark, which contains challenging high-definition, semantic, or text-rich visual queries.

\paragraph{Implementation} We implement \ourswb based on the LLM as Vicuna-7B-V1.3, and ViT 224/14, following LLaVa's architecture. We finetune our models on 8 Nvidia GPUs, with a learning rate of $2e-5$, batch size of 16, for one epoch, with a cosine scheduler and Adam optimizer. 

For pre-training, we use the 558K subset from LAIONCC-SBU, following LLaVA. Subsequently, we fine-tune on a 427K dataset, comprising 130K negative (for agent call generation) and 297K positive examples\footnote{The LLaVA 7B and 13B baselines in this work are also reproduced by fine-tuning on the 297K positive examples, following \citep{wu2023v}. The difference is that no extra clues from agents are provided for the 217K hard queries.}. Our negative data includes 110K objects from \citep{wu2023v} and 20K text images\footnote{Selected for their critical text dimensions < 20 pixels.} from DocVQA, ChartVQA, and TextVQA. The positive data consists of 80K simple questions from VQA train sets (for direct-answering training) and 217K challenging samples for agent utilization (190K object images from \citep{wu2023v} and 27K text images from Doc, Chart, and TextVQA).

\begin{table}[!tbp]
\tabcolsep 0.02 in
\centering
\resizebox{\linewidth}{!}{
\begin{tabular}{l|c|p{2.7cm}<{\centering} |p{2.7cm}<{\centering} |p{2.7cm}<{\centering}}
\toprule
Benchmark     & \ours & w/o Position in Prompt & w/ Weaker DINO & w/o Agents \\
\midrule
DocVQA        & 61.4    & 71.6 \textcolor{red}{(+10.2)}   & 61.4 \textcolor{gray}{(0.0)}  & 19.0 \textcolor{blue}{(-42.4)}   \\
ChartVQA      & 37.2    & 26.8 \textcolor{blue}{(-10.4)}    & 37.2 \textcolor{gray}{(0.0)}  & 15.3 \textcolor{blue}{(-21.9)}  \\
SEED          & 27.5    & 24.6 \textcolor{blue}{(-2.9)}     & 27.4 \textcolor{blue}{(-0.1)}  & 23.5 \textcolor{blue}{(-4.0)}    \\
MM-VET        & 32.9    & 29.1 \textcolor{blue}{(-3.8)}     & 29.3 \textcolor{blue}{(-3.6)}  &  29.1 \textcolor{blue}{(-3.8)}   \\

\bottomrule
\end{tabular}}
\caption{Effects on removing the relative position of grounded (text and/or visual) objects in prompt (\textit{w/o Position in Prompt}), replacing the visual grounding agent with a weaker, non-finetuned DINO (\textit{w/ Weaker DINO}), and removing agents in \ourswb (w/o Agents).}
\vspace{-10px}
\label{tab:aba}
\end{table}

\subsection{Results}
\subsubsection{Performance on Visual Reasoning}
The performance of \ourswb on visual reasoning benchmarks is presented in Table \ref{tab:main_tab}. On text-rich visual reasoning, \ourswb significantly outperform baselines, including the vanilla LLaVA, by more than doubled (3$\times$ on DocVQA, 2.4$\times$ on ChartVQA), and also greatly surpass MLLMs that dedicated tuned for text-rich visual reasoning, e.g., LLaVAR and TGDoc, and even surpasses 13B LLaVA variants. On general visual reasoning benchmarks, \ourswb also enjoys a consistent improvement over LLaVA and InstrtuctBLIP, demonstrating the superiority of \ours.

\subsubsection{Performance on \oursBenchmark}
On the more challenging \oursBenchmark, \ours achieved a significant improvement over LLaVA, demonstrating a markedly enhanced comprehension of object details in high-resolution images by over $5x$ compared with vanilla LLaVA. \ourswb is also comparable to GPT-4V and significantly outperforms baselines on reasoning related with nuanced \textit{Objects}, the most capable MLLM so far, and is huge in scale and training compute. These promising results further highlight the significance of \ourswb in plug-and-play grounding. A detailed case study on \oursBenchmarkwb against GPT-4V is illustrated in Figure \ref{fig:case}.

\subsubsection{Ablation Study} We study the effect of \ourswb in Table \ref{tab:aba}. We first remove the two agents for plug-and-play grounding (w/o Agents) by providing no additional clues, and the performance drops drastically, indicating the significance of Plug-and-Play Grounding. Upon removing the relative position vector for grounded objects and texts, we observed a performance degradation across multiple benchmarks. This decrement was more notable in structured image datasets like ChartVQA, where grounding bounding boxes are essential for the model to locate crucial text pieces\footnote{In DocVQA, we discover that removing bounding boxes unintentionally enables room for more detected texts within the maximized input token limitation (2K). We expect a positive effect of bounding boxes, given an MLLM with longer context.}. We finally replaced the grounding agent with a weaker model that not being fine-tuned\footnote{Both versions: \texttt{longzw1997/Open-GroundingDino}}. It drops improvements in benchmarks that require both object and text recognition, such as MM-VET, while it does not impact benchmarks focused solely on text recognition, like DocVQA.

\section{Analysis}
To further understand the role of deliberate reasoning in \ours, we present a comprehensive analysis of this capability in \ours, on SEED, which contains both text- and visual-rich samples \citep{li2023seed}.

\begin{table}[!ht]
\centering
\resizebox{\linewidth}{!}{
\begin{tabular}{c|c|p{1.5cm}<{\centering}p{1.3cm}<{\centering}p{1.3cm}}
\toprule
Model & Size & Simple & Hard \\
\midrule
LLaVA & 7B & 29.58 & 14.86 \\

LLaVA + \ourswb (Ours) & 7B & \textbf{33.67} & \textbf{18.57} \\

\midrule
Gain (\%) & - & \textcolor{blue}{13.8} & \textcolor{blue}{25.0} \\

\bottomrule
\end{tabular}}
\caption{Performance \ours and baselines under simple and hard questions in SEED.}
\vspace{-10px}
\label{tab:analysis_perf}
\end{table}

\paragraph{Performance Gain via Agent Assistance} We first study the effect of deliberate reasoning, under both \textit{simple} and \textit{hard} visual queries. To obtain such splits, we leverage a strong, larger model LLaVa-V1.5-13B. We treat the samples whose answers are correct as simple sets, and vice versa. As listed in Table \ref{tab:analysis_perf}, our \ours is able to improve performance on both easy and difficult tasks, while the improvement is greater for difficult topics. This suggests that our deliberate reasoning allows the model to answer simple questions more confidently while being able to use extrinsic agents to improve performance on complex questions.

\begin{figure}[h]
  \centering
  \includegraphics[width=0.45\textwidth]{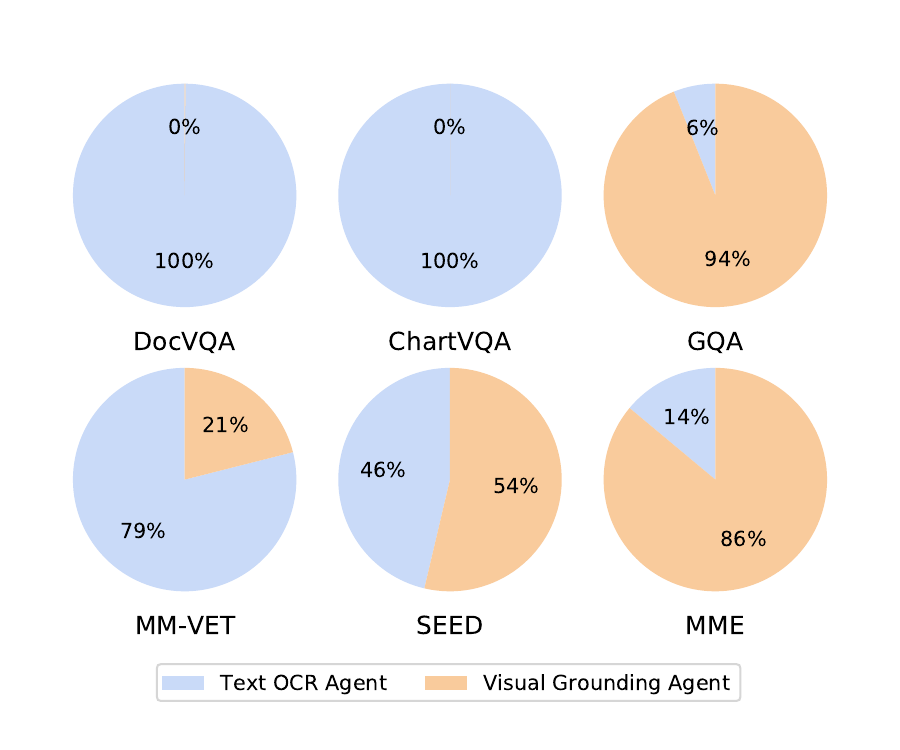}
  \caption{Agent routing of \ours under various tasks.}
  \label{fig:routing}
  \vspace{-5px}
\end{figure}

\paragraph{Routing to Different Agents} We further study the routing to each (OCR or Grounding) agent in \ours. As illustrated in Figure \ref{fig:routing}, both two types of agents are called during inference, indicating that \ours is capable of utilizing corresponding agents for reasoning in need (for text or visual clues).

\paragraph{Case Study of \ours}
We first perform a case study on \oursBenchmark, in Figure \ref{fig:case}, where we compare rationales generated by \ours and GPT-4V(ision). As illustrated in the figure, \ours could generate more grounded and accurate answers, especially for text-rich and high-resolution samples. To further understand the deliberate reasoning process of \ours, we provide detailed case studies in Appendix \ref{app:case}.

\section{Conclusion}
In this paper, we focus on the challenge of grounding visual reasoning of multimodal large language models. To address the limitations of most existing works that heavily rely on question-answer pairs for instruction tuning, we propose \ours, a novel framework for plug-and-play grounding of visual reasoning. Dedicated tuned to deliberate thinking, \ourswb promptly generates calls on external agents for detailed text and visual clues within the image, thus performing better reasoning. Furthermore, we propose \oursBenchmark, a challenging benchmark with text-rich and high-definition images to better assess reasoning capabilities. Comprehensive experiments on a variety of datasets demonstrate the superiority of \ours, especially under text-rich and high-definition images. Our work provides meaningful insights into the enhancement of MLLM reasoning capabilities with tool usage and plug-and-play grounding. We provide a detailed discussion on related works to \ours in Appendix \ref{sec:related}.

\newpage
\section*{Limitations}
In this section, we discuss the limitations of the current work in detail, outlining future directions.

1) Noise in agents. It is a shared common challenge on the capability of external agents itself \citep{liang2023taskmatrix, shen2024hugginggpt} in tool-augmented (M)LLMs. While we leverage state-of-the-art agents when building \ours, it is possible that it returns noisy, biased, or inaccurate results. In the future, we may propose a post-agent-call filtration strategy, or explore recent advances like self-consistency \citep{wang2022self}.

2) Token count. To incorporate finer multimodal semantics into contexts for grounded reasoning, \ours inevitably leverages a longer context of input. To accommodate more tokens, we propose novel routing strategies for MLP or resampler-based token compression mechanisms. However, we believe it is also promising to explore enhancing \ours with efficient sampling approaches, e.g. KV-Caching.

3) Modality-interleaved or multi-hop reasoning. Another limitation of current work and valuable future direction is to expand \ours into multi-hop and complex reasoning that involves interleaved multi-modality clues. For future studies, we may explore expanding types of agents, and adapting tree \citep{treetotnips} or graph-structured \citep{gotaaai} reasoning or agent calling paths for supporting these more challenging scenarios.

\section*{Ethnics Statement}
This work studies enhancing smaller MLLMs on visual reasoning via leveraging external agents and deliberate reasoning, which improves the reasoning capability of smaller MLLMs and potentially makes them more helpful by improving the accuracy and groundedness of their answers.

All visual images for creating our novel benchmark \oursBenchmark are from publicly accessible resources, which we adequately cited in our paper. On corresponding verbal multiple-choice questions, for ones we adapted from existing works, we cite their sources in our paper; and we leverage a publicly-accessible model (GPT-4o) to synthesize the rest and manually double-check their correctness. The proposed benchmark will be publicly released.

\bibliography{egbib}
\appendix
\begin{figure*}[!htbp]
  \centering
  \includegraphics[width=\textwidth]{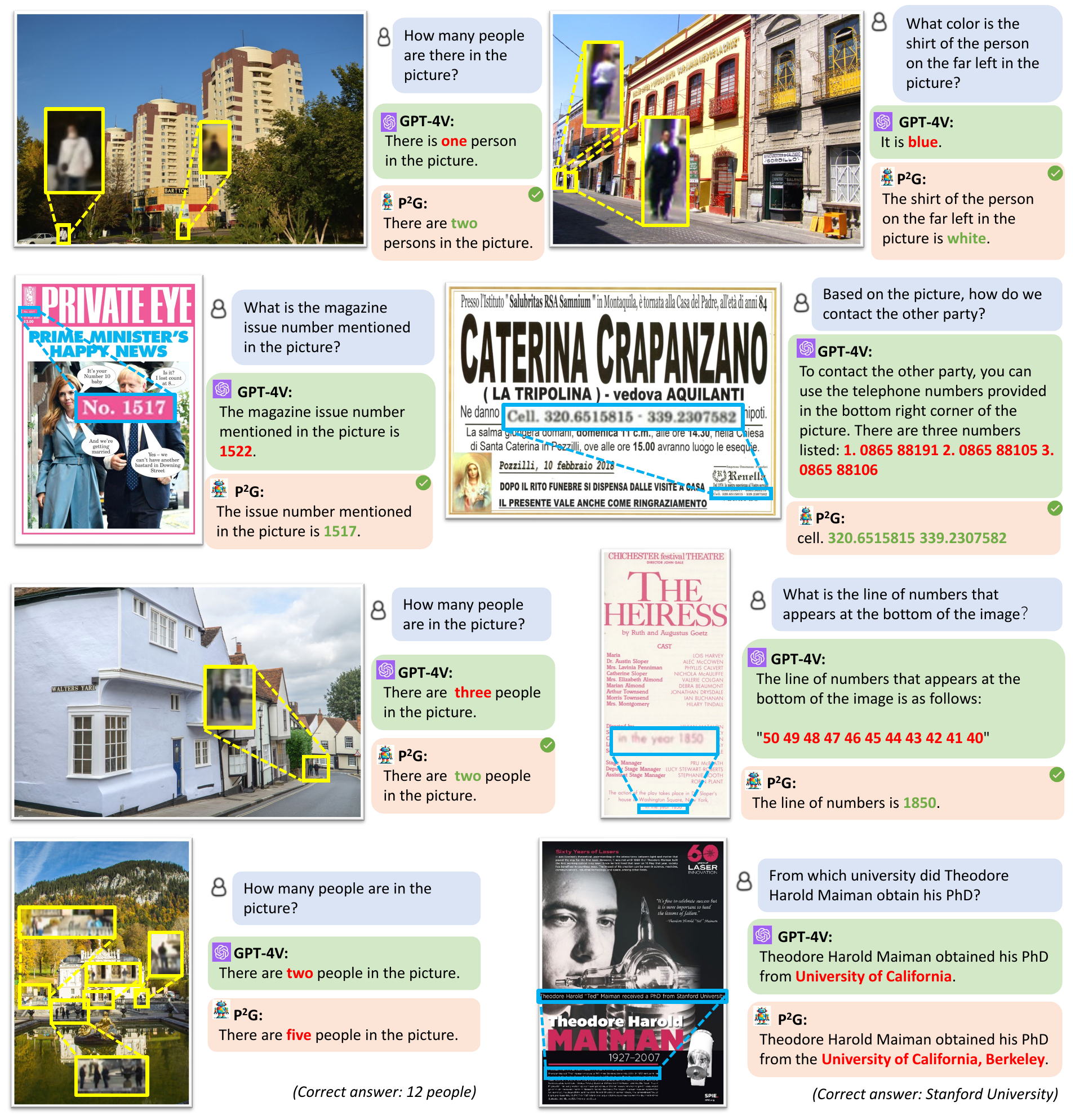}
  \vspace{5px}
  \caption{Case study of visual reasoning on \oursBenchmark, where we compare rationales generated by \ourswb and GPT-4V(ision). The first three lines from top to bottom demonstrate cases on both text-rich and semantic-rich reasoning, and bounding boxes generated with \textit{OCR agent} and/or \textit{Grounding Agent} of \ours, where \ourswb (based on LLaVA-7B) demonstrates its superior capability in generating grounded reasoning leveraging additional semantic clues against GPT-4V. The last row comprises two challenging failure cases where both \ourswb and GPT-4V fails in generating an accurate answer.}
  \label{fig:case}
\end{figure*}

\begin{table*}[!ht]

\centering
\resizebox{\linewidth}{!}{
\begin{tabular}{c|p{8cm}|p{8cm}}
\toprule
 & \textbf{Case \#1} & \textbf{Case \#2}  \\
\midrule
Question & What is the color of the bowl on the counter? A. Blue B. Green C. White D. Silver & Is there any musical instrument seen on the stage? A. No, there isn't. B. Yes, there is a drum. C. Yes, there is a guitar. D. Yes, there is a piano. \\

\midrule

Image Size & $3264 \times 2448$  & $2048 \times 1536$ \\

\midrule

Agent Returns & \includegraphics[width=8cm]{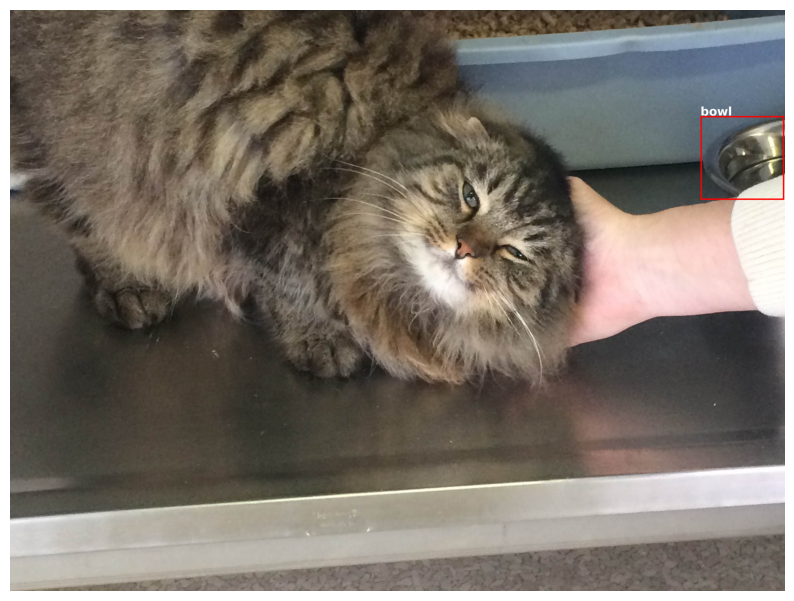} & \includegraphics[width=8cm]{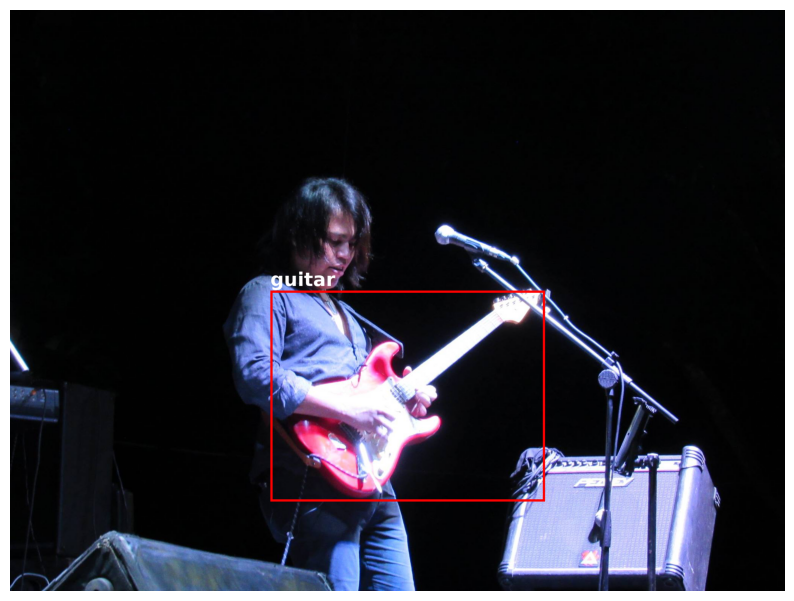} \\

\midrule 

Final Prompt & Additional visual information to focus on: 1 bowl <object> at location [0.891,0.184,0.999,0.328]

What is the color of the bowl on the counter? A. Blue B. Green C. White D. Silver 

Answer with the option's letter from the given choices directly. & Additional visual information to focus on: 1 guitar <object> at location [0.336,0.484,0.690,0.846] 

Is there any musical instrument seen on the stage? A. No, there isn't. B. Yes, there is a drum. C. Yes, there is a guitar. D. Yes, there is a piano. 

Answer with the option's letter from the given choices directly. \\
\midrule

Final Answer & \ours (Ours): \textcolor{cyan}{D} \quad LLaVa: \textcolor{red}{B} & \ours (Ours): \textcolor{cyan}{C} \quad LLaVa: \textcolor{red}{B} \\

\bottomrule
\end{tabular}
}
\caption{Two cases of Plug-and-Play grounding of \ours to critical objects in high-resolution images.}
\label{tab:case_deli_image}
\end{table*}


\begin{table*}[!ht]

\centering
\resizebox{\linewidth}{!}{
\begin{tabular}{c|p{8cm}|p{8cm}}
\toprule
 & \textbf{Case \#3} & \textbf{Case \#4}  \\
\midrule
Question & How would you describe the general appearance of the buildings in the photo? A. Modern and sleek B. Colorful and unique C. Industrial and metallic D. Old and brick & How much alcohol is in this beverage? \\

\midrule

Image Size & $736 \times 938$  & $550 \times 1200$ \\

\midrule

Agent Returns & \includegraphics[width=8cm]{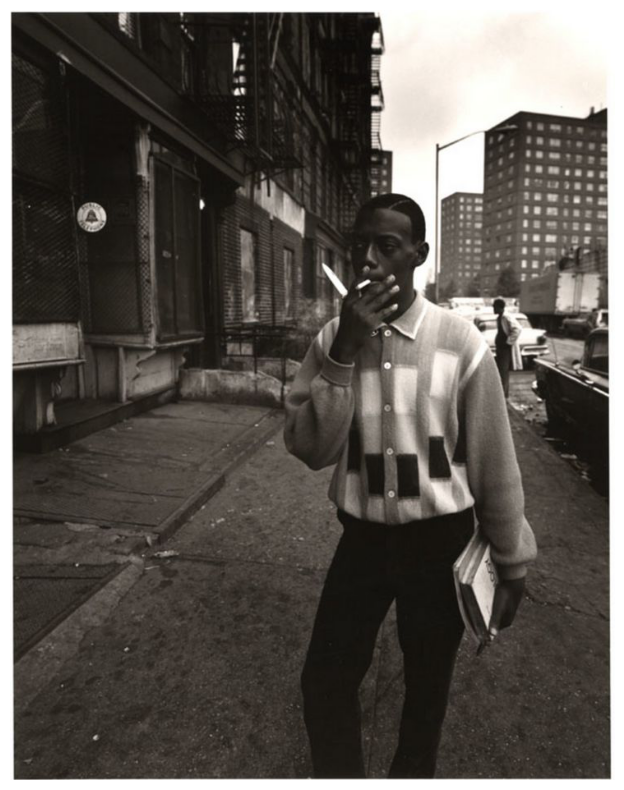} \textit{(no texts detected in the image)} & \includegraphics[width=8cm]{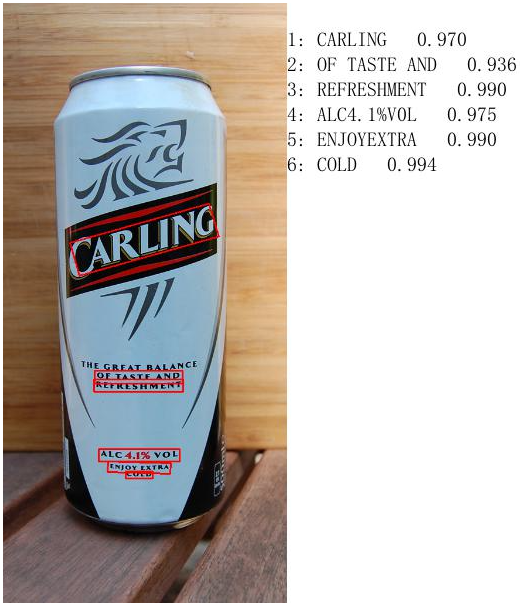} \\

\midrule 

Final Prompt & Additional visual information to focus on:

\textit{Please focus on providing an answer to the question without considering any challenges related to the clarity or presence of text in the image.}

How would you describe the general appearance of the buildings in the photo? A. Modern and sleek B. Colorful and unique C. Industrial and metallic D. Old and brick

Answer with the option's letter from the given choices directly. 

\textit{(no texts detected in the image)}
& Additional visual information to focus on: text in the image: 

`CARLING' at location [0.227, 0.333, 0.757, 0.454]; 

`OFTASTE AND' at location [0.330, 0.614, 0.623, 0.629]; 

`ALC4.1\%VOL' at location [0.340, 0.743, 0.619, 0.764]; 

`ENJOY EXTRA' at location [0.373, 0.767, 0.588, 0.781]; 

`COLD' at location [0.433, 0.780, 0.522, 0.791]

How much alcohol is in this beverage? \\
\midrule

Final Answer & \ours (Ours): \textcolor{cyan}{D} \quad LLaVa: \textcolor{red}{A} & \ours (Ours): \textcolor{cyan}{4.1\%} \quad LLaVa: \textcolor{red}{2\%} \\

\bottomrule
\end{tabular}}
\vspace{5px}
\caption{Two cases of Plug-and-Play grounding of \ours to critical texts that tiny in its scale. \textit{Left}: when no texts are detected by OCR agent, we inform the model and encourage it to focus on non-textual semantics. \textit{Right}: when critical texts are detected, we incorporate them with their relative position in multimodal query.}
\label{tab:case_deli_text}
\end{table*}

\section{Related Works}
\label{sec:related}
\subsection{Multimodal LLMs}
The surge of large language models (LLMs) \cite{Achiam2023GPT4TR, touvron2023llama}, especially instruction-tuned ones \cite{longpre2023flan, chiang2023vicuna, touvron2023llama2, mukherjee2023orca} demonstrated a strong potential in becoming generic interface for language modality. To extend LLMs beyond language perception, recent works \cite{zhu2023minigpt, liu2024visual-llava1, huang2024language, alayrac2022flamingo, wang2023cogvlm, instructblip} extends them into Multimodal Large Language Models (MLLMs) with instruction tuning, through incorporating each modality as a foreign language\cite{huang2024language, wu2023next}. To equip LLM with capability in image perception, pioneer works like Flamingo \cite{alayrac2022flamingo} and BLIP-2 \cite{li2023blip} first encode image with a dedicated model (e.g.ViT \cite{dosovitskiy2020vit}), then propose specific modules for aligning image and text modality. Subsequent works like LLaVA \cite{liu2024visual-llava1} and \textsc{Kosmos-1} \cite{huang2024language} leverage vision tokenizers to feed image semantics as in-context tokens, thereby aligns the perception of image and language. To further advance MLLMs, recent works explored enabling grounding and reference to visual contexts \cite{peng2023kosmos, wang2023cogvlm}, generating contents leveraging multimodal adaptors \cite{wu2023next, pan2023kosmosg}, leveraging parameter-efficient fine-tuning \cite{gao2023llama, shen2024multimodal}, and scaling of multimodal instruction data and model parameters \cite{Achiam2023GPT4TR, bai2023qwen, lu2024deepseek}. Despite these improvements, MLLMs so far still suffers from multiple prevailing limitations, including high-demand on quality and quantity of instruction-following data, hallucination \cite{liu2024survey}, and difficulties in processing images within text-rich contexts \cite{wang2023towardstgdoc} or grasping details within high-resolution images \cite{liu2024visual-llava1}.

\subsection{Visual Reasoning in Text-Rich Images}
\citet{zhang2023llavar} developed LLaVAR, which aims to enhance the interactive capabilities of MLLMs through improved visual instruction tuning for text-rich image understanding. \citet{hu2023bliva} introduce BLIVA, which employs a novel approach by integrating both learned query embeddings and image-encoded patch embeddings to enhance the multimodal LLM's understanding and processing of text-rich visual questions. \citet{wang2023towardstgdoc} focus on enhancing MLLMs with text-grounding to improve document understanding, especially in text-rich scenarios. Despite employing extensive instruction fine-tuning data, the models' capability for text grounding remains limited. \citet{wadhawan2024contextual} emphasize the need for models to understand interactions between text and visual content in their evaluation of context-sensitive text-rich visual reasoning in large multimodal models. They primarily employ OCR tools and GPT-4 to construct instruction-finetuned datasets that enhance MLLM's visual reasoning of text-rich images; however, mere instruction fine-tuning struggles to effectively leverage LLM's potent generative capabilities, resulting in marginal improvements.

\section{Extended Case Study}
\label{app:case}
To further understand plug-and-play grounding of reasoning in \ours, we provide two case studies in Table \ref{tab:case_deli_image} and \ref{tab:case_deli_text}. As illustrated in Table \ref{tab:case_deli_image}, \ours could effectively utilize additional visual clues from Grounding Agent to improve its accuracy of answers, compared to LLaVA. As illustrated in Table \ref{tab:case_deli_text}, by providing textual clues from OCR Agent, the capability of \ours in understanding tiny texts are also largely improved. These cases further highlights the effectiveness of \ours's design.
\end{document}